\documentclass[conference]{IEEEtran}
\IEEEoverridecommandlockouts
\usepackage{cite}
\usepackage{amsmath,amssymb,amsfonts}
\usepackage{algorithmic}
\usepackage{graphicx}
\usepackage{textcomp}
\usepackage{multirow}
\usepackage{flushend}
\usepackage{enumitem}
\usepackage{graphicx}
\usepackage{booktabs}
\usepackage[table]{xcolor}

\setlist[itemize]{noitemsep, leftmargin=*, topsep=0pt}
\def\BibTeX{{\rm B\kern-.05em{\sc i\kern-.025em b}\kern-.08em
    T\kern-.1667em\lower.7ex\hbox{E}\kern-.125emX}}
\begin{document}

\title{AquaNeRF: Neural Radiance Fields in Underwater Media with Distractor Removal
\thanks{This work is supported by EPSRC ECR (EP/Y002490/1) and UKRI MyWorld Strength in Places Programme (SIPF00006/1)}
}

\author{Luca Gough, Adrian Azzarelli, Fan Zhang, Nantheera Anantrasirichai \\

\IEEEauthorblockA{\textit{Visual Information Laboratory}, \textit{University of Bristol}, Bristol, UK \\
}

}

\maketitle

\begin{abstract}
Neural radiance field (NeRF) research has made significant progress in modeling static video content captured in the wild. However, current models and rendering processes rarely consider scenes captured underwater, which are useful for studying and filming ocean life. They fail to address visual artifacts unique to underwater scenes, such as moving fish and suspended particles. This paper introduces a novel NeRF renderer and optimization scheme for an implicit MLP-based NeRF model. Our renderer reduces the influence of floaters and moving objects that interfere with static objects of interest by estimating a single surface per ray. We use a Gaussian weight function with a small offset to ensure that the transmittance of the surrounding media remains constant. Additionally, we enhance our model with a depth-based scaling function to upscale gradients for near-camera volumes. Overall, our method outperforms the baseline Nerfacto by approximately 7.5\% and SeaThru-NeRF by 6.2\% in terms of PSNR. Subjective evaluation also shows a significant reduction of artifacts while preserving details of static targets and background compared to the state of the arts.
\end{abstract}


\section{Introduction}
Underwater filming is costly and limited by experts and technology. Advanced imaging techniques now allow remote exploration, enabling experts such as geologists and archaeologists to model underwater environments.  Three-dimensional (3D) reconstructions from image sequences enhance understanding of underwater organisms, objects, damage, and seabed.

Neural radiance fields (NeRFs) model objects, scenes, and people in 3D using posed images or video frames. Since the introduction of vanilla NeRF \cite{mildenhall2021nerf}, research has expanded to include applications such as aerial photography and large-scale modeling \cite{dronova2024flynerf, zhang2024birdnerf, jia2024drone} and casually captured content \cite{martin2021nerf, kaneko2022ar, chen2022hallucinated}. However, the use of NeRF in dense media remains underexplored. 
The earliest attempt, SeaThru-NeRF \cite{levy2023seathru}, incorporates unique per-ray parameters like backscatter, attenuation density, and medium color using a separate MLP. These parameters depend on ray direction and disentangle media properties (transmittance and color) from the object's characteristics, separating them for enhanced clarity. WaterNeRF \cite{Sethuraman:WaterNeRF:2023} estimates parameters from a physics-based underwater light transport model \cite{Schechner:Recovery:2005} and performs color correction via optimal transport for consistent color across viewpoints. Both methods perform well for static scenes but struggle with temporal occlusions, such as fish or particles. UWNeRF \cite{10656460} models static and dynamic objects separately. The results heavily rely on the estimated mask, and the rendered dynamic objects sometimes appear more like artifacts than real objects.

\begin{figure}[t!]
    \centering
   \includegraphics[width=\linewidth]{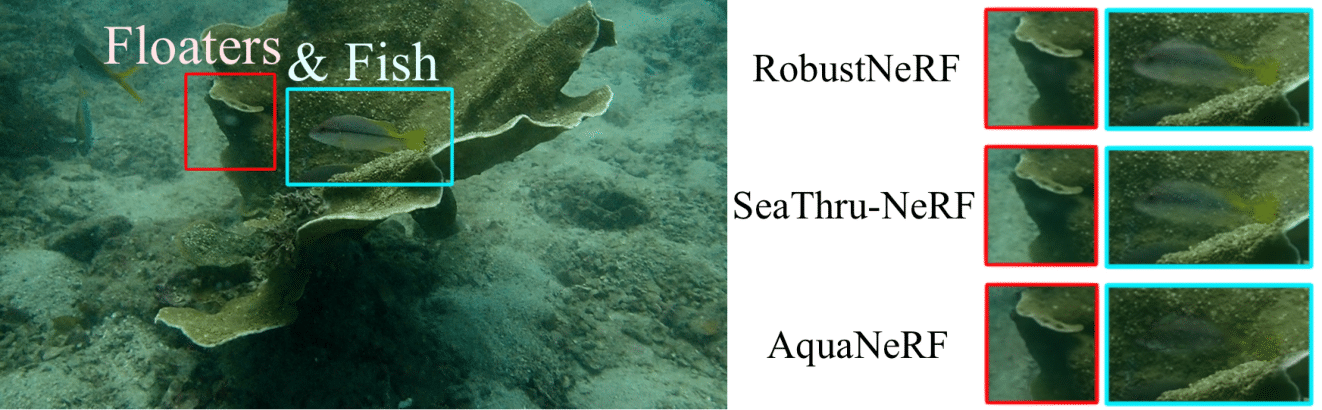}
   \vspace{-20px}
    \caption{{Challenges of modeling underwater scenes:} (Red) View-dependent suspended particles and (Blue) Dynamic Fish.}
    \label{fig:teaser}
\end{figure}

The key challenges in modeling underwater scenes \cite{levy2023seathru}: (1) light-medium interaction causing view-dependent chromatic aberration; (2) dynamic, view-dependent floating particulates near the camera; and (3) visibility issues at greater depths, resembling low-light problems \cite{marques2020l2uwe, liu2021hazing}. Our research adds that lens distortion, motion blur, and interactions between dynamic objects (e.g., fish) and the static scene also pose challenges. These interactions can cause blur-like effects, vividly shown in Fig.~\ref{fig:teaser}, where rendered frames from RobustNeRF \cite{sabour2023robustnerf} and SeaThru-Nerf \cite{levy2023seathru} depict semi-transparent fish and significant floater artifacts, contrasting with our proposed technique's results. 

In this paper, we address the above challenges by introducing a new renderer that models the cumulative density of volumes along a ray and redistributes transmittance from low-density regions to high-density regions. We propose using a single Gaussian distribution to model the new transmittance function, which confines the in-frustum geometry to model regions around a single surface. This results in sampling higher density volumes at the target during optimization, leading to a more robust interpretation of the scene's geometry. This approach also mitigates issues with view-dependent high-density suspended particles, as the proposal sampler is encouraged to generalize geometry to a single surface across all views.

Additionally, we integrate a method for gradient scaling during optimization. Gradient scaling can be useful for prioritizing the learning of specific features. In our case, we prioritize features that appear further away from a camera by upscaling gradients proportional to distance. This works under two assumptions: (1) suspended particles are more visible closer to the camera, and (2) distant (unimportant) background objects are less visible, so they are less likely to discourage learning depth if we prioritize objects far from the camera in denser media.

In summary, the contributions outlined are as follows:
\begin{enumerate}
\item Introduction of a new renderer that models the cumulative density of volumes along a ray and redistributes transmittance from low-density regions to high-density regions. 
\item Utilization of a single Gaussian distribution to model the transmittance function, confining the in-frustum geometry to model regions around a single surface.
\item Integration of gradient scaling during optimization to prioritize learning to focus on features that are potentially more relevant in dense media environments.
\end{enumerate}


\section{Methodology}
\label{sec:model}

\begin{figure}[t!]
    \centering
    \includegraphics[width=\linewidth]{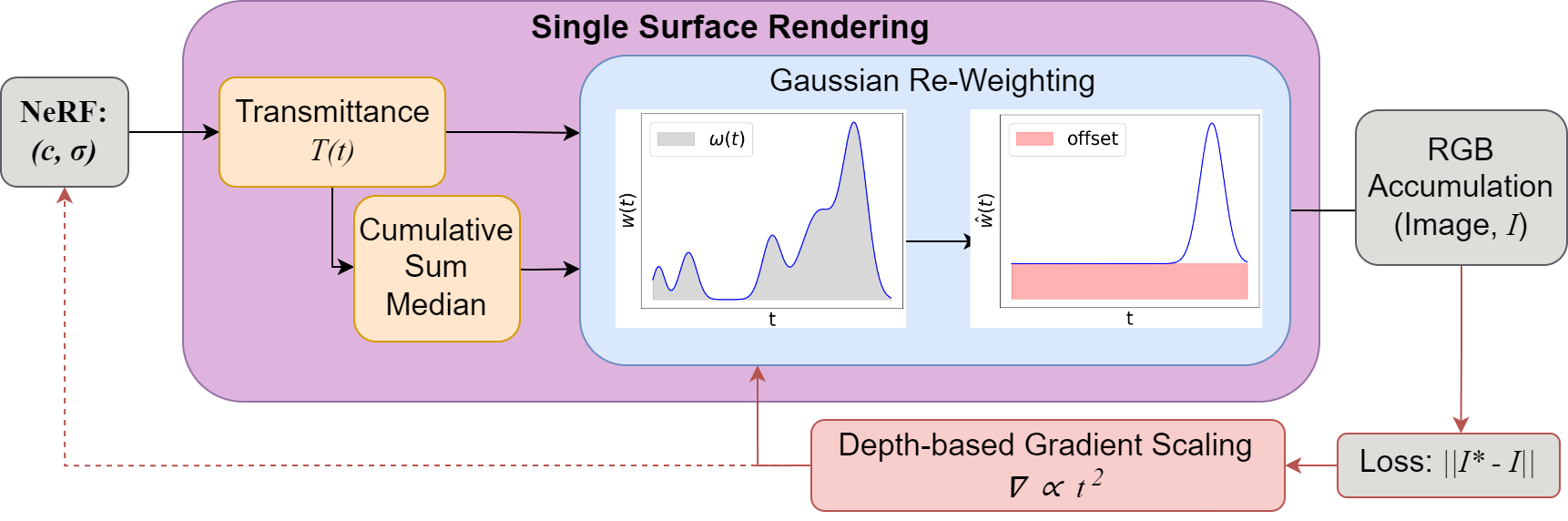}
    \caption{\textbf{Proposed Rendering Overview:} Using the color and density outputs, we model the weight distribution $w(t)$ and re-formulate the distribution of weights with respect to $\hat{w}(t)$. }
    \label{fig:model}
\end{figure}

An overview of our contributions is illustrated in Fig.~\ref{fig:model} with respect to the NeRF pipeline \cite{mildenhall2021nerf}. We reinterprete the rendering equation proposed in \cite{Max:Optical:1995},
\begin{equation}\label{eq:original.rendering.equation}
    C(\mathbf{r}) = \int_{t_{n}}^{t_{f}} T(t)\cdot \sigma(\mathbf{r}(t)) \cdot \mathbf{c} ( \mathbf{r}(t), \mathbf{d})\cdot dt,
\end{equation}
where the color of a pixel $C(\mathbf{r})$ (associated with the ray $\mathbf{r}(t)$ and direction $\mathbf{d}$) is recovered by modeling the transmittance $T(t)$, the density $\sigma(\cdot)$ and color $\mathbf{c}(\cdot)$ of volumes sampled between $t_{n}$ and $t_{f}$ along the camera ray $\mathbf{r}(\cdot)$. We address limitations with using the accumulated transmittance function,
\begin{equation}\label{eq:origina;.transmittance.equation}
    T(t)= exp ( - \int_{t_{n}}^{t} \sigma(\mathbf{r}(u)) \cdot du),
\end{equation}
to define the weights, 
\begin{equation}
    w(t) = T(t) \cdot (1 - exp(-\mathbf{\sigma}(\mathbf{r}(t)) \cdot \delta(t))),
\end{equation}
where $\delta(t)$ is the length occupied by a volume along a ray and $w(t)$ replaces $T(t)\cdot \sigma (\cdot)$ in Eq.~\ref{eq:original.rendering.equation}, resulting $C(\mathbf{r}) = \int_{t_{n}}^{t_{f}} w(t) \cdot \mathbf{c} ( \mathbf{r}(t), \mathbf{d})\cdot dt$. 
This relies on the assumption that, as light scatters in denser media the transmittance of light from an opaque object to a camera diminishes as the distance between the two increases. This has volumetric significance in Eq.~\ref{eq:origina;.transmittance.equation} and results in a wider distribution, as shown in Fig.~\ref{fig:model}. In our case, we define a compact distribution of transmittance by re-distributing the weights, $w(t)$, around a single point, assumed to be the surface of the target object. This avoids the need to model volumes containing dense media (found in-between the camera and target object) as was done in SeaThru-NeRF, so reduces the complexity of our approach in comparison.

\subsection{Single Surface Rendering -- AquaNeRF}
The existing methods model the target object and dense media as separate trainable parameters. This invites additional complexity and leads to instability in the presence of distractors.
Similarly, we disentangle the object from the environment. Though, our description of the environment is not limited to just the dense media but everything except the static target object. We accomplish this by only modeling a single surface per ray, using the original prediction for $T(t)$ to inform the estimation of depth for a single volume. This is illustrated by Fig.~\ref{fig:model} comparing the initial (left) and final (right) distributions of weights.

We define the depth at which a ray-surface collision occurs as the median point, which is used as the mean $\mu_t$. This is the point, $\mu_t = t$, at which the cumulative sum of weights,
\begin{equation}\label{eq:cumsum.nerfstudio}
    \omega (t) = \int_{t_n}^{t} w(u) du,
\end{equation}
crosses the threshold $\omega (t) > 0.5$. We then reduce the new distribution of weights to a Gaussian with the mean $\mu_t$ and fixed standard deviation $\eta$, using
\begin{equation}\label{eq:GaussianDistribution}
    \hat{w}(t)=\frac{1}{\sqrt{2\pi}\eta}exp(\frac{-(t-\mu_t)^2}{2\eta ^2}).
\end{equation}

We found that the $\eta$ values of 0.3-0.6 give the best results in terms of PSNR, so we set $\eta$ to 0.5 for the rest of this paper. The Gaussian distribution is specially useful as the area enclosed under the function $\hat{w}(t)$ is 1, derived in Eq.~\ref{eq:GaussianIntegral}. This is consistent with the integral of the original weight distribution $w(t)$. The resulting transmittance is subsequently redistributed normally around the single surface point without modifying the distribution of color along a ray.
\begin{equation}\label{eq:GaussianIntegral}
    \int_{-\infty}^{\infty}w(t)dt=\lim_{r\rightarrow\infty}\left( \sum_{t=t_n}^{t_f}\frac{w(t/r)}{r}\right) = 1.
\end{equation}

As the derivative $\delta w(t)$ is equivalent to $w(t)$, the gradients of the volumes samples follow the same distribution. During optimization, this places significantly more attention on learning the color and position of the single surface Gaussian distribution; allowing for more stable optimization without involving additional learnable parameters.

Finally, to model the dense media we apply an offset to $\hat{w}(t)$ to increase the minimum density value associated with non-surface volumes. This ensures the weights associated with the media are fixed and constant, meaning the media is modeled as low-density and evenly distributed throughout the scene. The maximum value is kept at $\frac{1}{\sqrt{2\pi}\eta}$. Therefore, the final weights is defined as
\begin{equation}
    \hat{w}(t) =  \min(\frac{1}{\sqrt{2\pi}\eta},\hat{w}(t) + \text{base}).
    \label{eqn:finalweight}
\end{equation}

 In this paper, we set the value of base to 0.2. It is observed that the novel view with a small offset value exhibits better detail compared to the view without it.

\begin{figure*}
    \centering
    \includegraphics[width=1.\linewidth]{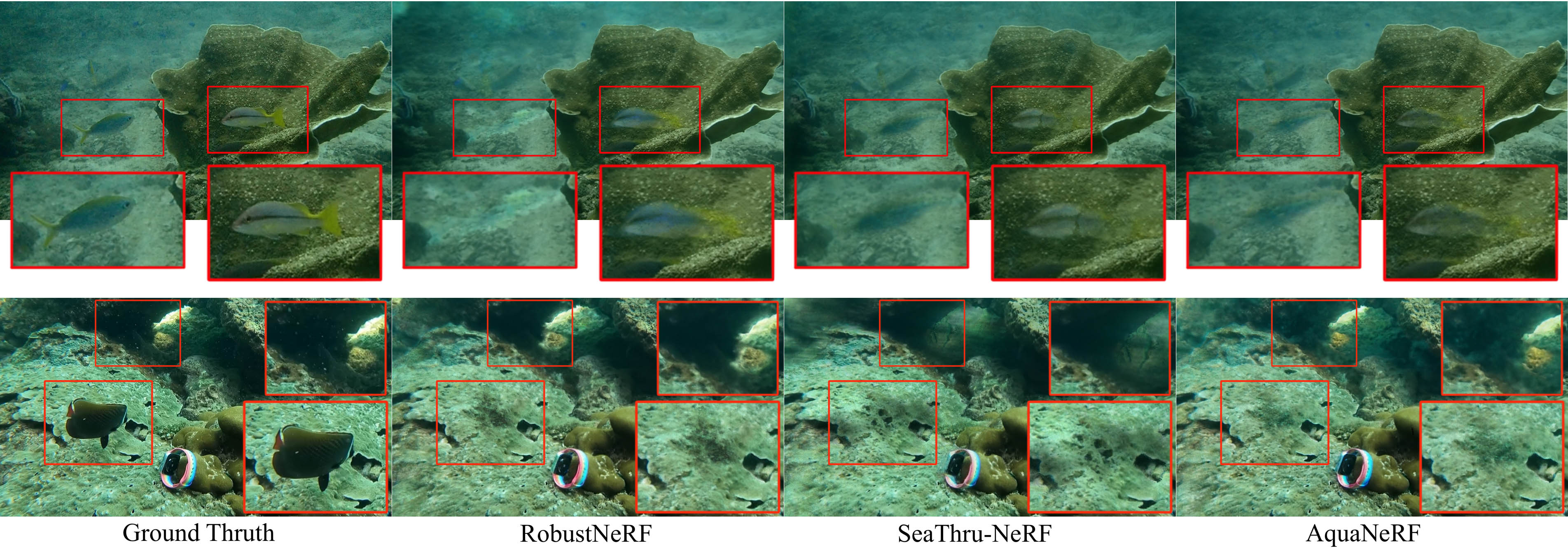}
    \caption{{Visual comparison} of rendered results from the BVI-Coral dataset. The top row contains suspended particles, and both rows contain dynamic fish.}
    \label{fig:fullcomparison}
\end{figure*}

\subsection{Robust Loss and Depth-based Gradient Scaling}
Following \cite{sabour2023robustnerf}, we include a Robust Loss (RL) to decrease the model's sensitivity to outliers, which is done by applying a 3$\times$3 smoothing kernel, $\mathcal{B}_{3\times3}$ to an $\mathcal{L}_2$ RGB loss:
\begin{equation}
\mathcal{L}_{\text{rgb}} = ||C_{GT}(\mathbf{r}) - C(\mathbf{r})||^2_2, 
\end{equation}
where $GT$ is the ground truth using  $\mathcal{W}(r)=(\mathcal{L}_{\text{rgb}}^r*\mathcal{B}_{3\times3})\ge 0.5.$
Then, an 8$\times$8 neighborhood of pixels is classified as outliers or inliers, based on a convolution across a larger 16$\times$16 patch, leveraging the spatial locality of outliers as described as $
    \Upsilon(R_8(r))=A_{t\in R_{16}(r)}(\hat{w}(t)) \ge T_R$,
where $T_R = 0.6$ represents the threshold, and $A$ is an aggregation function over the specified neighborhood.

Finally, the $\mathcal{L}_2$ loss is re-weighted using Iteratively Reweighted Least Squares (IRLS), as described in Eq.~\ref{eq:Robust_Weighting}.
\begin{equation} 
\mathcal{L}_{\text{robust}}^r=\Upsilon(R_{8}(r))\cdot \mathcal{L}_{\text{rgb}},
    \label{eq:Robust_Weighting}
\end{equation}
 where $R_N(r)$ defines an $N \times N$ neighborhood around $\mathbf{r}$. This method effectively adjusts the weighting of the loss based on the localized evaluation of outlier presence.

In practice, this approach is challenging to fine-tune. We hence enforce gradient scaling only when the average depth $D_{avg}$ of a given view is above a defined threshold $T_h$. This avoids under-fitting near-frustum objects, while allowing gradient scaling for images in the training set that contain a higher proportion of distant observations. Eq.~\ref{eq:Gradient_Scaled_Losses} denotes our Depth-based Gradient Scaling (D-GS) applied to the RGB loss, $\partial C$, with respect to a particular weight $\partial w_i^{(L)}$, for a sample at position $p$ from a camera positioned at $c$.
\begin{equation}
    \nabla_{\text{grad}}= 
\begin{cases}
    \frac{\partial C}{\partial w_i} \cdot \min (1, \left |c - p\right |^2),& \text{if } \frac{1}{M} \sum_{i=0}^M (\left|c - p_i\right|) > T_h\\
    \frac{\partial C}{\partial w_i},              & \text{otherwise}
\end{cases}
\label{eq:Gradient_Scaled_Losses}
\end{equation}
 where $M$ is the number of rays. Empirically, $T_h$ is set to 0.02. 

\subsection{Implementation and Optimization}
We implement our approach on the Nerfstudio platform using a three-layer Multilayer Perceptron (MLP) with 128 neurons to estimate color and density. For volumetric sampling, we adopt the proposal sampling scheme from MipNeRF360 \cite{multinerf2022}, prioritizing high-density regions. Early stopping is performed every 2000 iterations by monitoring Peak Signal-to-Noise Ratio (PSNR), halting training if PSNR declines. Instead of the Adam optimizer, we use Rectified Adam (RAdam) for more stable convergence with a learning rate of 0.01 \cite{liu2021variance}. Pose estimation is performed using the COLMAP software \cite{schoenberger2016sfm}.

\section{Results and Discussion}
\label{sec:results}
Performance is evaluated by PSNR, SSIM, and LPIPS, through visual comparisons. We compared our method against SeaThruNeRF \cite{levy2023seathru}, RobustNeRF \cite{sabour2023robustnerf}, and used Nerfacto \cite{nerfstudio} as a baseline. WaterNeRF \cite{Sethuraman:WaterNeRF:2023} was excluded due to its reliance on raw underwater images. 
We utilized the BVI-Coral \cite{BVI-Coral2024} and S-UW \cite{wang2025uwgs} datasets to evaluate our models. These scenes feature varying degrees of turbidity from low to medium and include challenges such as noise, small floating particles, and occasional interference from animals. 
We divided each video into training, validation, and testing segments with an 80:10:10 split, respectively.

\subsection{Performance Comparison}

Rendering static areas from an image sequence with moving distractors is mainly assessed through visualization, as objective evaluation requires ground truth. In our application, this relies on static scenes without distractors. Though this does not fully capture the paper's main objective, it demonstrates that the proposed framework should maintain performance in distractor-free areas.

\subsubsection{Scenes with moving distractors}
Fig.~\ref{fig:teaser} and \ref{fig:fullcomparison} demonstrate the effectiveness of AquaNeRF in scenes filled with suspended particles and fish, showing a significant reduction in scene distractors. For example, in the bottom row of Fig.~\ref{fig:fullcomparison}, AquaNeRF notably diminishes the visibility of dark suspended particles and the associated shadow effects that are prominent in the SeaThruNeRF render. We also tested our method on the clear medium. As shown in Fig.~\ref{fig:Non-subsea_Dynamic_Object_Removal}, AquaNeRF deals with distractors better than Nerfacto.

\subsubsection{Scenes without moving distractors}
Although AquaNeRF is designed for dynamic objects and suspended particles in underwater media, we evaluated its performance on static scenes. Table~\ref{tab:averaged_performance_v123} presents metrics from the frames without dynamic objects for accurate comparison with ground truth.  AquaNeRF surpasses Nerfacto and SeaThru-NeRF, showing improvements of 7.5\% and 6.2\%, respectively. While AquaNeRF matches RobustNeRF in static scenes, RobustNeRF struggles with artifacts from dynamic objects, as seen in Fig.~\ref{fig:teaser} and Fig.~\ref{fig:fullcomparison}. This is likely due to RobustNeRF's focus on minimizing obstructions from distant viewpoints, resulting in clearer backgrounds but less effective handling of closer, larger artifacts. Despite similar SSIM and LPIPS metrics, AquaNeRF excels at handling moving distractors and performs comparably or better in static scenes.

\begin{table}[t!]
    \caption{Averaged performance of the scenes without moving objects. Highlighting in green indicates the best performances.} \vspace{-2mm}
    \label{tab:averaged_performance_v123}
    \begin{center}
        \begin{tabular}{cccc}
            \toprule
            Model &  PSNR  $\uparrow$ &  SSIM  $\uparrow$ &  LPIPS  $\downarrow$ \\
            \midrule
            Nerfacto \cite{nerfstudio}  & 23.23 & 0.647 & 0.270 \\
            SeaThru-NeRF \cite{levy2023seathru}  & 24.66 & 0.703 & 0.263 \\
            RobustNeRF \cite{sabour2023robustnerf} & 24.55 & \cellcolor{green!25}0.737 & \cellcolor{green!25}0.230 \\
            GS \cite{philip2023floaters} & 24.87 & 0.730 & \cellcolor{green!25}0.230 \\ \midrule
            AquaNeRF & \cellcolor{green!25}24.98 & 0.730 & \cellcolor{green!25}0.230 \\
            AquaNeRF+RL & 24.33 & 0.730 & 0.233 \\
            AquaNeRF+D-GS & 24.82 & 0.733 & 0.237 \\
            AquaNeRF+RL+D-GS & 24.28 & 0.727 & 0.240 \\
            \bottomrule
        \end{tabular}
    \end{center}
\vspace{2mm}
    \caption{{Gradient scaling} (GS) and {depth-based gradient scaling} (D-GS) comparison. Green indicates best performances.}
    \label{tbl:Gradient_Scaling_Table}
\centering
       \begin{tabular}{ccccc}
        \toprule
            & \multicolumn{2}{c}{PSNR $\uparrow$} & \multicolumn{2}{c}{LPIPS $\downarrow$} \\ \cline{2-3} \cline{4-5}
            Model & Video 1 & Video 2 & Video 1 & Video 2  \\ \midrule
            Nerfacto & 24.10 & 21.26 & 0.26 & 0.28 \\       
            Ours & 24.79 & \cellcolor{green!25}21.45 & \cellcolor{green!25}0.25 & \cellcolor{green!25}0.26 \\       
            Ours + GS & 24.74 & 21.12 & 0.27 & 0.27 \\       
            Ours + D-GS & \cellcolor{green!25}24.83 & 21.30 & 0.26 & \cellcolor{green!25}0.26\\      \bottomrule
        \end{tabular} 
  
\end{table}

\subsection{Ablations}
\subsubsection{Gradient Scaling and Robust Losses}
Table~\ref{tab:averaged_performance_v123} shows the impact of gradient scaling schemes on reducing floating artifacts. Our method achieves strong performance across a variety of scenes without the need for additional constraints. We observe that in scenes with fewer suspended particles, the depth-oriented solution is the most effective approach.

\begin{figure}
\begin{tabular}{@{}c@{}c@{}c@{}c@{}}
    \includegraphics[width=0.33\linewidth]{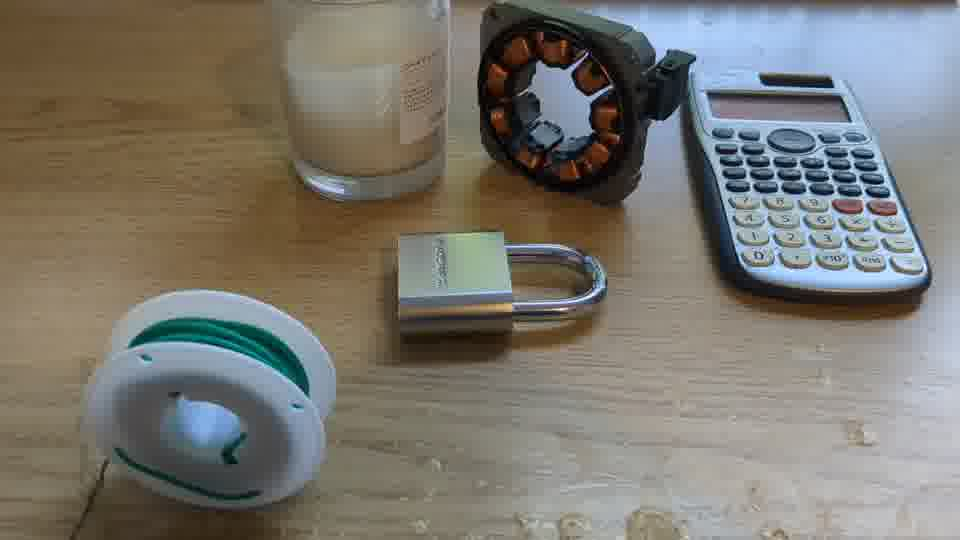} &     
        \includegraphics[width=0.33\linewidth]{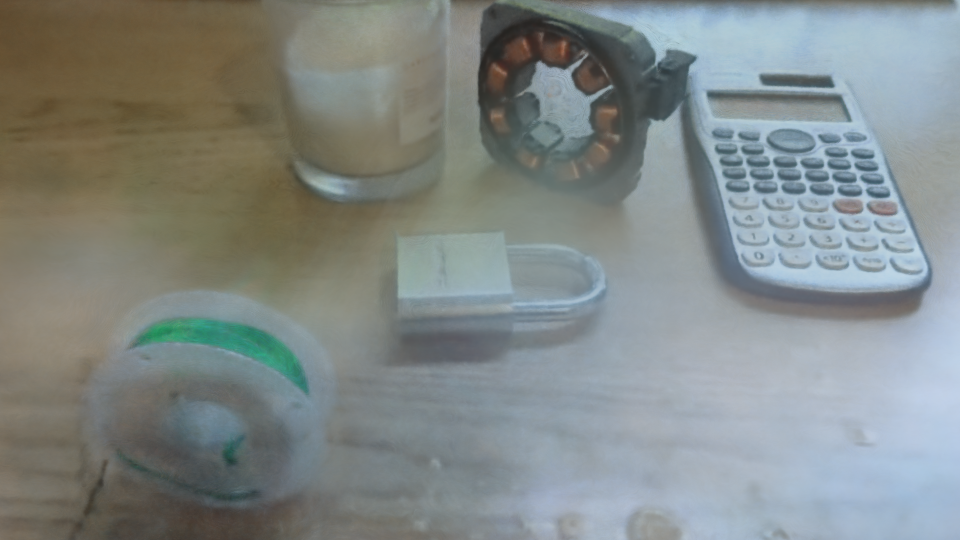} &
        \includegraphics[width=0.33\linewidth]{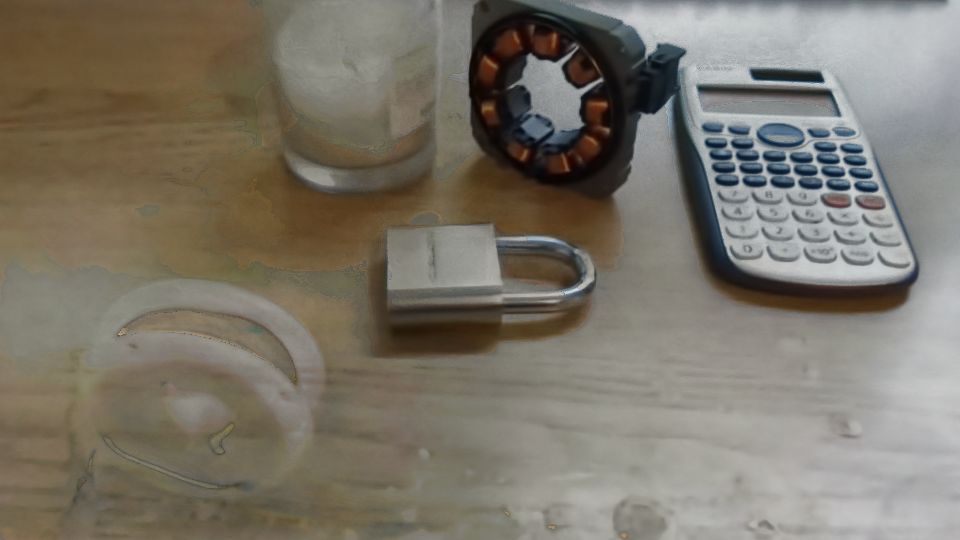} \\
    {\footnotesize Ground Truth} & {\footnotesize Nerfacto} & {\footnotesize AquaNeRF}
    \end{tabular}
    
    \caption{Dynamic object removal in clear medium}
    \label{fig:Non-subsea_Dynamic_Object_Removal}
\vspace{2mm}
    \centering
    \begin{tabular}{@{}c@{}c@{}}
    \includegraphics[width=0.5\linewidth]{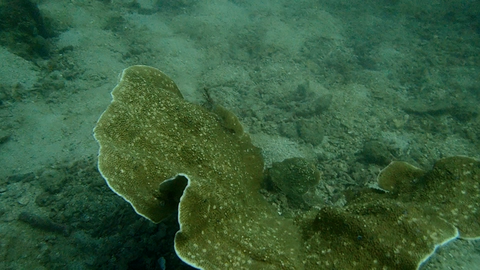} &
    \includegraphics[width=0.5\linewidth]{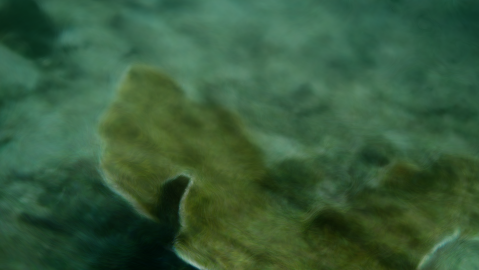} \\
    {\footnotesize Ground Truth} & {\footnotesize K-Planes}
    \end{tabular}
    \caption{{Visual comparison} of AquaNeRF rendering applied to K-Planes \cite{fridovich2023k}. The average PSNR across all scenes is 22.5.}
    \label{fig:k-planes}
\vspace{2mm}
    \centering
        \includegraphics[trim={0 25mm 0 0}, clip, width=\columnwidth]{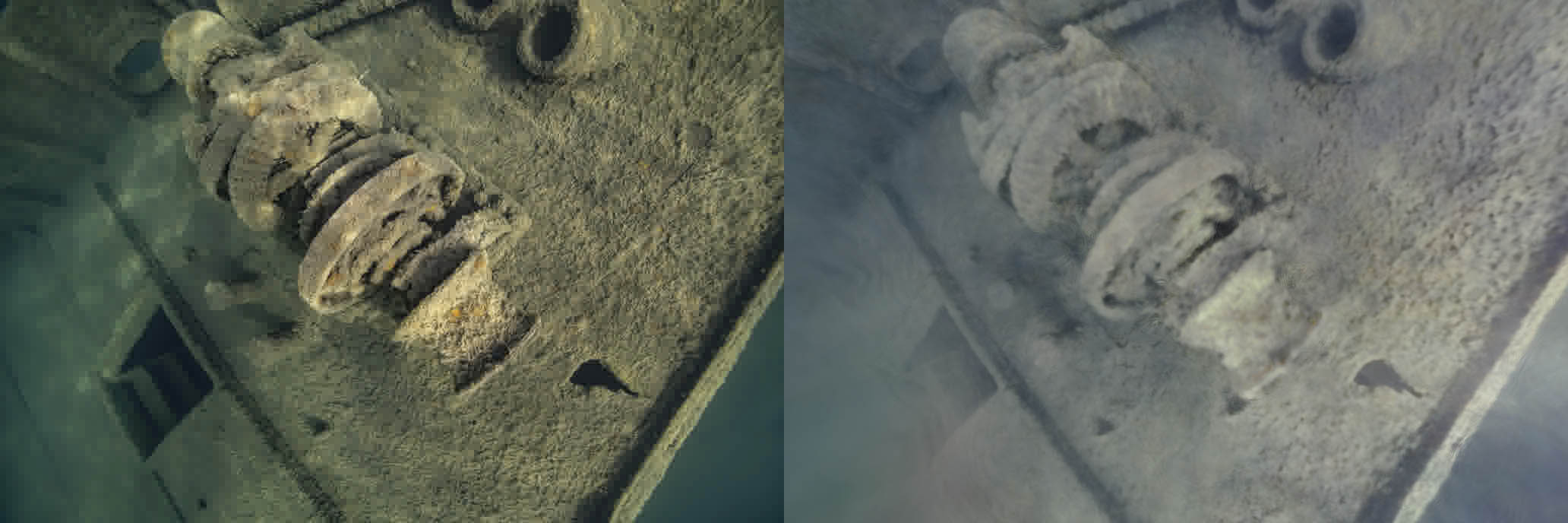}
    \caption{Shipwreck dataset evaluation. Left: the original dataset images. Right: the images generated by our model. 
    }
    \label{fig:Shipwreck_Eval_Images}
\end{figure}

\subsubsection{Dynamic NeRF} 
We implemented the AquaNeRF rendering technique on K-Planes \cite{azzarelli2023waveplanes}, a dynamic NeRF model designed to separate dynamic and static elements within a scene. Fig.~\ref{fig:k-planes} illustrates the visual outcomes of this experiment, revealing that K-Planes is less effective in capturing high-frequency details than the static version of AquaNeRF.

\subsection{Limitations}
When capturing still images at varying distances, our method struggled as rays from distant cameras were attenuated before reaching surface voxels. Assuming uniform density across angles led to low-density volume portrayal, effectively replacing water. As shown in Fig.~\ref{fig:Shipwreck_Eval_Images}\footnote{Thanks to the NPS Submerged Resources Center for providing dataset},the reconstructed images, though blurred and less color accurate, largely preserve the original structures from the dataset.

The training times for RobustNeRF, SeaThru-NeRF, and AquaNeRF are about 3 hours on the BVI-Coral dataset using a RTX3090 GPU, compared to Nerfacto, requiring about 2 hours. The method may expedite color and density inference processes with explicit representations \cite{chen2022tensorf, fridovich2022plenoxels, muller2022instant}.

\section{Conclusion}
\label{sec:conclusion}
This paper presents a novel approach for modeling static scenes in dense media using a re-weighting strategy with a slightly offset Gaussian distribution. This method effectively addresses floating and dynamic artifacts in underwater datasets. AquaNeRF improves both objective and visual performance over state-of-the-art methods with reduced complexity and simpler tuning. We explored various optimization schemes, ultimately selecting a depth-guided gradient scaling strategy that outperforms previous robust losses and gradient scaling methods. AquaNeRF, tailored to these challenges, offers more stable optimization compared to alternative schemes, which proved less reliable in practice.

\bibliographystyle{IEEEtran}
\bibliography{egbib}

\end{document}